\documentclass[nohyperref]{article}

\usepackage{microtype}
\usepackage{graphicx}
\usepackage{subfigure}
\usepackage{booktabs} 
\usepackage[all=normal, mathspacing=tight, floats=tight, paragraphs=tight]{savetrees}
\usepackage{hyperref}

\usepackage[accepted]{icml2023}

\usepackage{amsmath}
\usepackage{amssymb}
\usepackage{mathtools}
\usepackage{amsthm}

\usepackage[capitalize,noabbrev]{cleveref}

\theoremstyle{plain}

\theoremstyle{definition}

\theoremstyle{remark}

\usepackage[textsize=tiny]{todonotes}

\icmltitlerunning{Genomic Interpreter: A Hierarchical Genomic Deep Neural Network with 1-D Swin Transformer}

\begin{document}

\twocolumn[
\icmltitle{Genomic Interpreter: A Hierarchical Genomic Deep Neural Network\\
 with 1D Shifted Window Transformer}


\icmlsetsymbol{equal}{*}
\begin{icmlauthorlist}
\icmlauthor{Zehui Li}{be}
\icmlauthor{Akashaditya Das}{be}
\icmlauthor{William A V Beardall}{be}
\icmlauthor{Yiren Zhao}{eee}
\icmlauthor{Guy-Bart Stan}{be}
\end{icmlauthorlist}

\icmlaffiliation{be}{Department of Bioengineering, Imperial College London}
\icmlaffiliation{eee}{Department of Electrical and Electronic Engineering, Imperial College London}

\icmlcorrespondingauthor{Guy-Bart Stan}{g.stan@imperial.ac.uk}

\icmlkeywords{Machine Learning, ICML}

\vskip 0.3in
]



\printAffiliationsAndNotice{} 
\begin{abstract}
Given the increasing volume and quality of genomics data, extracting new insights requires interpretable machine-learning models. This work presents Genomic Interpreter: a novel architecture for genomic assay prediction. This model outperforms the state-of-the-art models for genomic assay prediction tasks. Our model can identify hierarchical dependencies in genomic sites. This is achieved through the integration of 1D-Swin, a novel Transformer-based block designed by us for modelling long-range hierarchical data. Evaluated on a dataset containing 38,171 DNA segments of 17K base pairs, Genomic Interpreter demonstrates superior performance in chromatin accessibility and gene expression prediction and unmasks the underlying 'syntax' of gene regulation.\footnote{We make our source code for 1D-Swin publicly available at \url{https://github.com/Zehui127/1d-swin}}
\end{abstract}

\section{Introduction}
Functional genomics uses a variety of assays to explore the roles of genes in a genome. These assays allow researchers to quantify gene expression \cite{de2008deep}, test chromatin accessibility, and understand gene regulation \cite{eraslan2019deep}. 
In this paper:

\begin{itemize}
    \item We introduce Genomic Interpreter, an attention-based model for genomic assay prediction.
    \item We design a task-agnostic hierarchical Transformer, 1D-Swin, for capturing long-range interaction in 1-D sequences.
    \item We show that our model performs better than the state-of-the-art.
    \item We further demonstrate that the hierarchical attention mechanism in Genomic Interpreter provides us with interpretability. This can help biologists to identify and validate relationships between different genomic sequences.
    
\end{itemize}

\section{Related Work}
\begin{figure*}[ht]
\begin{center}
\centerline{\includegraphics[width=\textwidth]{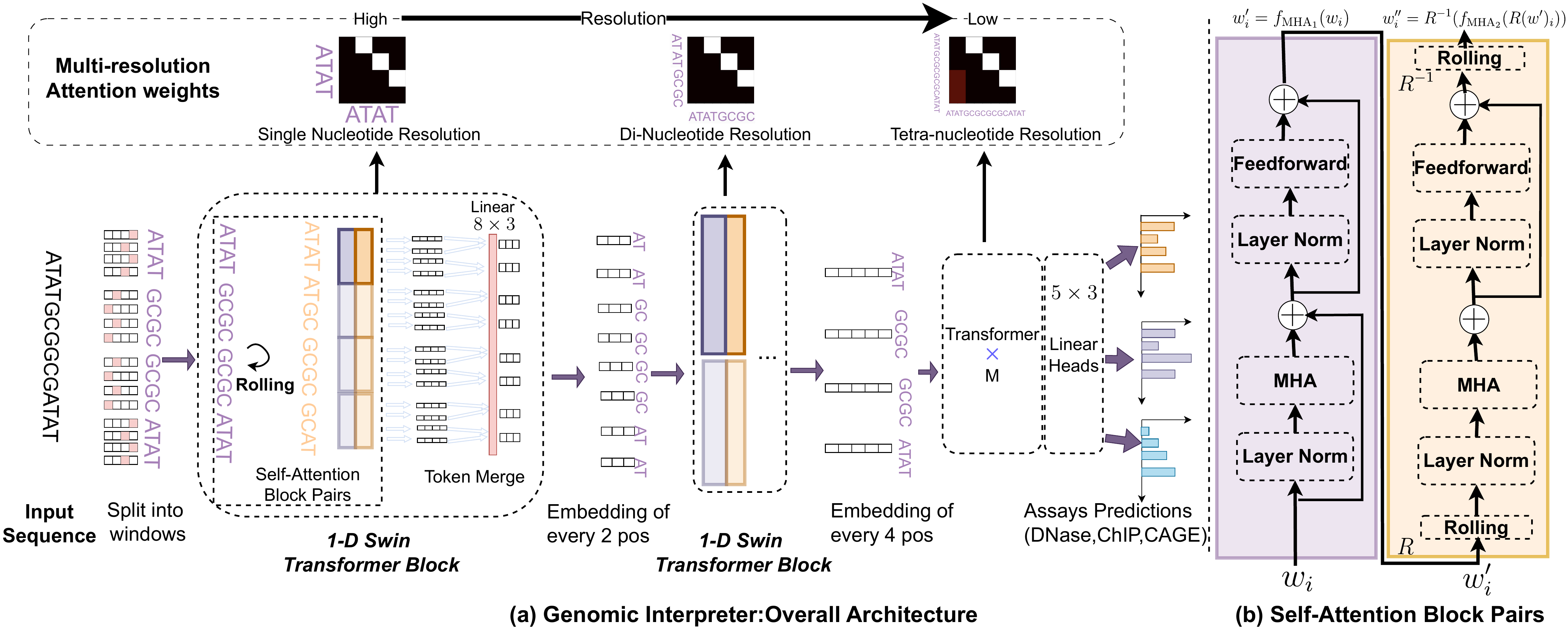}}
\caption{(a) Genomic Interpreter: given an input sequence $x \in \mathbb{R}^{16 \times 4}$, the target is to predict $3\times4$ read value arrays. $x$ first traverses multiple 1D-Swin blocks. Each forward pass halves the token length; after two passes, the token number is  reduced to 4. Then the resulting embedding is fed into Transformer Block and Linear Heads for final prediction. (b)Self-Attention Block Pairs: standard implementation is used for self-attention block}
\label{fig:architecture}
\end{center}
\vskip -0.2in
\end{figure*}

\textbf{Deep Learning for genomic assays prediction}
 Deep Neural Networks (DNNs) have seen success in predicting genome-scale sequencing assays such as CAGE 
 \cite{takahashi2012cage}, DNase-seq \cite{he2014refined}, and ChIP-seq \cite{zhou2017genome}. In genomic assay prediction, models are provided with a DNA sequence $x \in \mathbb{R}^{n\times4}$ and are required to forecast sequencing assay outputs $y \in \mathbb{R}^{m\times T}$ for various tracks, where $n$ defines the DNA sequence length, $4$ refers to four nucleotides in DNA, $m$ represents the length of the output sequence, with each output value being an average read over a specific DNA segment, and $T$ denotes the track count, referring to the different types of assay outputs being predicted, such as specific sequencing assays performed in particular organisms. In these models, DNA sequences are encoded and transformed into high-dimensional vector representations. These encoded vectors are then processed through a series of linear layers to predict the corresponding real-value assay readings. Existing genomic models can be divided into two categories: the first category uses Convolution Neural Networks (CNNs) and pooling layers as the encoder \cite{alipanahi2015predicting,kelley2018sequential,kelley2020cross}; the second type of model consists of CNNs and pooling layers followed by Transformer blocks, which is first proposed by \cite{avsec2021effective} as Enformer, serving as the state-of-the-art model used in regulatory DNA study \cite{vaishnav2022evolution}.


Current state-of-the-art models predict coarse-grained read values, typically over 100 base pairs per read. This requires DNNs to reduce the spatial dimension and create a condensed representation for the input. This property is also shared by most computer vision tasks such as object detection \cite{girshick2015fast} and semantic segmentation \cite{he2017mask}. While using CNNs and pooling layers has proven effective in the past. More recently, transformer-based architectures are becoming more popular and are outperforming CNN-based architectures \cite{dosovitskiy2020image,liu2021swin}. The success of these models motivated us to design attention-based models for genomic assay prediction. Furthermore, by calculating the Transformers' attention weights we can reveal genomic site dependencies \cite{ghotra2021uncovering}, thus providing us with better interpretability compared to CNN-based approaches in assay prediction.

\textbf{Swin Transformers} The Shifted Window (Swin) Transformer builds hierarchical feature maps of input images through self-attention operations within a local window and shift operations \cite{liu2021swin}. The computed feature maps are merged to higher layers, reducing the spatial size of inputs. Swin Transformer is efficient in processing long input sequences as self-attention is only computed within each window.

While Swin Transformer and its variants have shown improved performance in the computer vision tasks \cite{liu2022swin,yuan2021florence}. There is a lack of hierarchical transformer models for 1-dimensional data such as genomic sequence data. The local window approach has been adopted for long-range 1-D Transformer works such as  Longformer \cite{beltagy2020longformer} and Sparse Transformer \cite{child2019generating} to reduce the computation. However, no work has been done for reducing the spatial dimension in a hierarchical manner. This property is  crucial for tasks requiring spatial reduction, such as genomic assay prediction.

\section{Method}
\subsection{Model Architecture Overview}

We propose a novel model architecture, termed Genomic Interpreter, which is designed to predict genomic assays. The Genomic Interpreter is made up of several parts: multiple 1-dimensional Swin (1D-Swin) blocks, a transformer block \cite{vaswani2017attention}, and linear heads at the end. The structure is shown in Figure \ref{fig:architecture}(a). 

The process begins with an input sequence which acts as the initial token. This is  
denoted as $x \in \mathbb{R}^{n \times d}$, where $n$ is the token length and $d$ is the dimension for each token. One round of the 1D-Swin block transforms this matrix where $f_{\text{1dSwin}} : \mathbb{R}^{n \times d} \rightarrow \mathbb{R}^{\frac{n}{2} \times \frac{2d}{\alpha_0}}$, 
where $\alpha_0$ is the hidden size scaling factor. When $\alpha_0$ is set to 1, the hidden size of the token is doubled.

The transformed matrix is denoted as  $\mathbf{h}$. The number of 1D-Swin blocks (represented by K) is chosen to match the output length to the initial token length. If an exact match is not possible, a Crop operation is used to adjust to output length by removing certain elements from the ends of $\mathbf{h}$.

The overall architecture of the Genomic Interpreter can be summarized by three equations:
\begin{equation}
    \mathbf{h}_K = f_{\text{1dSwin}}(\mathbf{h}_{K-1}),  \quad \mathbf{h}_0 = \mathbf{x}   \label{eq:equation1} 
\end{equation}
\begin{equation}
    \mathbf{h_c} = f_{\text{crop}}(\mathbf{h_K})  \label{eq:equation2} 
\end{equation}
\begin{equation}
    \mathbf{y}  = \text{LinearHeads}(\text{TransformerBlock}(\mathbf{h_c})) \label{eq:equation3}
\end{equation}

Here, $\mathbf{h^K}$ represents the output after K iterations of the 1D-Swin transformation, with dimensions $ \frac{n}{2^K} \times \frac{2d}{(\alpha_0\alpha_1...\alpha_K)}$. This output is then passed through the Transformer blocks and linear heads in a feedforward fashion, yielding a per-track prediction denoted as $\mathbf{y}$. The dimensions of $\mathbf{y}$ are $m \times T$. As an example, $m=4$ and $T=3$ in Figure \ref{fig:architecture}(a).

\subsection{1D-Swin Block}
The standard Transformer model's quadratic time-space complexity, denoted as $\Omega(n^2)$ for input tokens $x \in \mathbb{R}^{n \times d}$, hinders its efficiency with long inputs like DNA sequences. The 1D-Swin Transformer reduces this complexity\footnote{The time complexity of 1D-Swin depends upon a hyperparameter: the window size. It ranges from $O(n)$ for a window size of 1 to $O(n^2)$ for a window size of $n$} to $\Omega(n)$ through two identical Multi-Head Attention (MHA) blocks. 

 The first Multi-Head Attention (MHA block), as depicted in Figure \ref{fig:architecture}(b), processes each a subset of tokens with a window to capture local dependencies. And then a rolling operation together with the second MHA block is applied to capture the cross-window dependencies. Finally, pair-wise concatenation and linear transformation are used to form the spatially reduced token set. For the rigorous definition of this process, please refer to Appendix \ref{appendix:a}.

\subsection{Multi-resolution genome sites dependency detection}
Capturing the interactions between sub-sequences in the genome is crucial for genomics science. Stacked 1D-Swin blocks allow  Genomic Interpreter to have hierarchical representations of such genomic sequences. We can look at the learned attention patterns to see how tokens are interacting with each other. 
 
 Specifically, as shown at the top of Figure-\ref{fig:architecture}(a), the self-attention scores between each token are extracted at each layer, serving as a fine-grained map showing how nucleotide sequences of different lengths interact with each other. 

\subsection{Data}
Understanding the dependencies between genomic sites requires a comprehensive dataset. The dataset provided by Enformer  \cite{avsec2021effective} offers a broad spectrum of gene expression data. However, it demands an unrealistic amount of computational resources to replicate their  algorithm training\footnote{On original dataset, the training time of full-size Enformer model requires 64$\times$3 TPU days} for practical usage. To address this issue, we have developed a scaled-down version of genomic datasets named `BasenjiSmall'. 

The `BasenjiSmall' dataset comprises 38,171 data points $(X, Y)$. This is the same amount of data points as the original Enformer data set. Each data point consists of a DNA segment $\mathbf{x}$ and assay reads $\mathbf{y}$. $\mathbf{x} \in \mathbb{R}^{17,712 \times 4}$ represents a DNA segment spanning $17,712$ base pairs (bp). $\mathbf{y} \in \mathbb{R}^{80 \times 5313}$ refers to the coarse-grained read values across $80 \times 128$ bp and 5313 tracks expanding various cell and assay types, these tracks include:
(1) 675 DNase/ATAC tracks, (2) 4001 ChIP Histone and ChIP Transcription Factor tracks and (3) 639 CAGE tracks.

This reduced dataset maintains the richness of gene expression data but is more manageable in terms of computational requirements. 
\subsection{Training}
We utilized Pytorch Lightning framework with (Distributed Data Parallelism) DDP strategy for parallel training.

All the models shown in Section \ref{section:results} are implemented in Pytorch and trained with 3 NVIDIA A100-PCIE-40GB, with a training time of 3*10 GPU hours per model.  The batch size is set to 8 for each GPU, with an effective batch size of 24. Adam optimizer \cite{kingma2014adam} is used together with the learning rate scheduler, CosineAnnealingLR \cite{gotmare2018closer} and a learning rate of 0.0003 to minimize the Poisson regression loss function for all models.

\section{Results}
\label{section:results}
\begin{table}[t]
\caption{Pearson correlation of 5 models for genomic assay prediction on 5313 tracks, tracks are grouped into DNase, CHIP histone/transcription factor, and CAGE.}
\label{table:main}
\begin{center}
\begin{small}
\begin{sc}
\begin{tabular}{lcccr}
\toprule
 Model Name        & DNase & ChIP & CAGE & Overall\\
\midrule
 MaxPoolCNNs   & 0.4996 & 0.4320 & 0.2581 &0.4195 \\
 AttPoolCNNs  & 0.4970 & 0.4346 & 0.2041  &0.4146\\
AttPoolDilate & 0.4778 & 0.4395 & 0.1803  &0.4130\\
Enformer      & 0.5462 & 0.4575&  \textbf{0.3307} & 0.4536\\
1D-Swin       & \textbf{0.5583} & \textbf{0.4670} &   0.3242 & \textbf{0.4614}\\
\bottomrule
\end{tabular}
\end{sc}
\end{small}
\end{center}
\vskip -0.2in
\end{table}

\begin{figure*}[ht]
\begin{center}
\centerline{\includegraphics[width=0.8\textwidth]{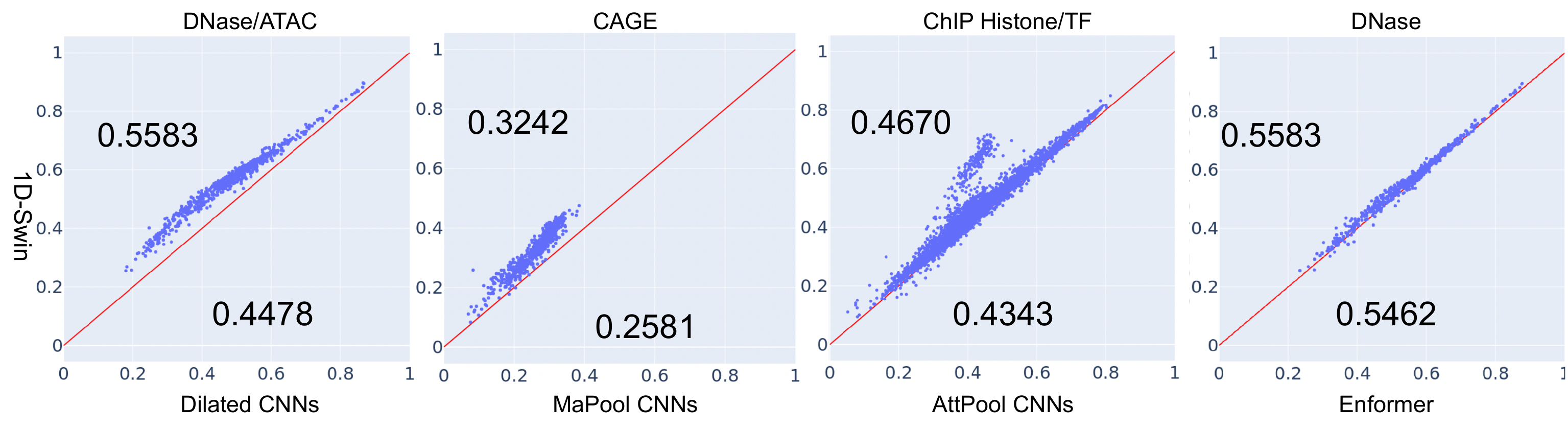}}
\caption{Pairwise Comparison between 1D-Swin and Competing Models: The Pearson correlation for each track is calculated across all DNA segments within the test set. Mean Pearson correlations are denoted on Y-axis for 1D-Swin and X-axis for the reference models.}
\label{fig:figure2}
\end{center}
\vskip -0.2in
\end{figure*}

\begin{figure}[ht]
\vskip 0.01in
\begin{center}
\centerline{\includegraphics[width=\columnwidth]{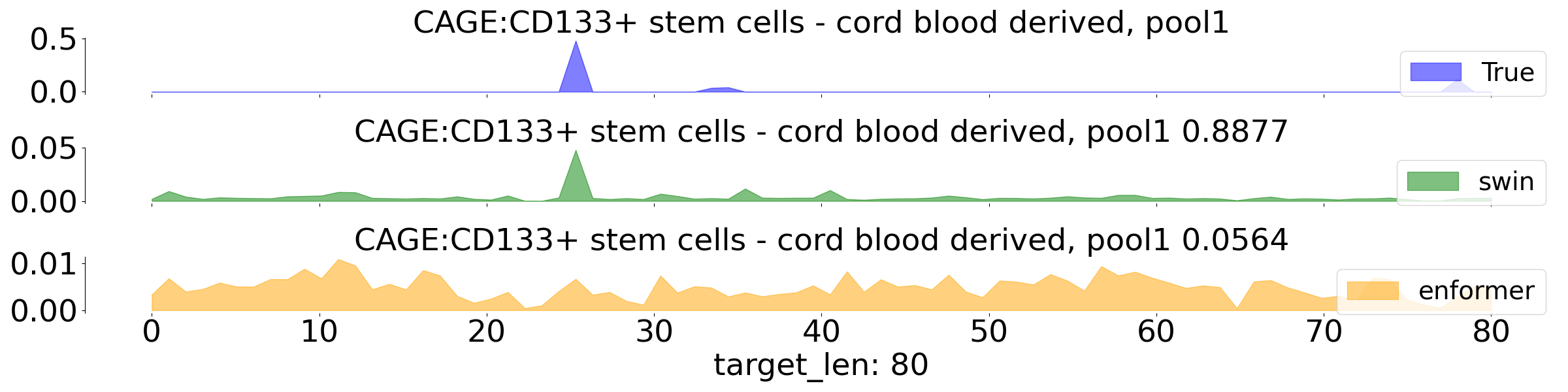}}
\caption{Given a DNA segment, two models will make predictions for the read values of the CAGE-CD133 track.}
\label{fig:figure3}
\end{center}
\vskip -0.2in
\end{figure}

\begin{figure}[ht]
\vskip 0.1in
\begin{center}
\centerline{\includegraphics[width=\columnwidth]{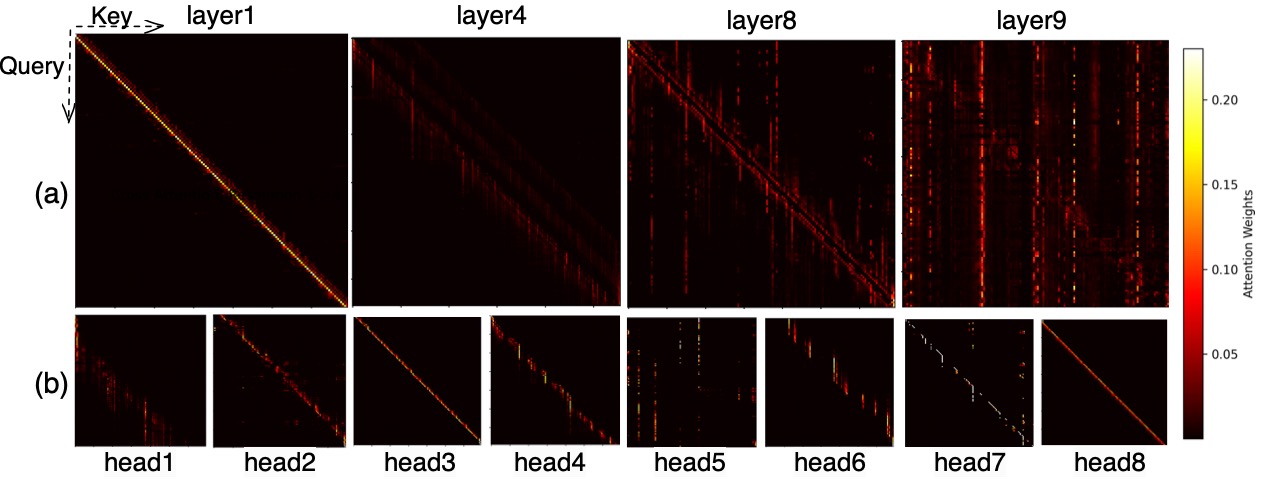}}
\caption{(a) Attention weights from different 1D-swin layers reveal a pattern to have more off-diagonal dependency at Higher Hierarchy (b) Attention weights of different heads on the 8th layer learn to capture various features.}
\label{fig:figure4}
\end{center}
\vskip -0.2in
\end{figure}

\subsection{Gene Expression Prediction} 

We compared the performance of 1D-Swin with Enformer \cite{avsec2021effective} and other standard genomic assay prediction models using a hold-out test set from BasenjiSmall. This included models implementing CNNs with MaxPooling \cite{alipanahi2015predicting}, CNNs with AttentionPooling, and Dilated CNNs with Attention Pooling \cite{kelley2018sequential}. For a detailed view of the implementation process, refer to Appendix \ref{appendix:b}.

Model performance is evaluated using the Pearson correlation between predictions and true values. Table \ref{table:main} shows the evaluation results for five models across 1937 DNA segments with 5313 tracks, classified into DNase, ChIP, and CAGE groups. These results show that 1D-Swin outperforms other models overall, particularly in the DNase and ChIP groups.

Figure \ref{fig:figure2} shows a pairwise comparison of 1D-Swin with other methods across track groups. Points above the reference line indicate superior performance by 1D-Swin. Figure \ref{fig:figure3} visualize the gene expression predictions from two models for CD133 stem cells, the visualisation evident that the prediction provided by 1D-swin matches more with the true values.


\subsection{Self-Attention and Gene Regulation}

While attention does not equate to explanation \cite{pruthi2019learning}, it helps to provide insights into token interaction. Genomic Interpreter leverages this capability to extract hierarchical self-attention values at varying resolutions, potentially unmasking the underlying 'syntax' of gene regulation.

In this hierarchical structure, each matrix position represents a single nucleotide at the first layer and $2^{K-1}$ nucleotides by the Kth layer. Figure \ref{fig:figure4}(a) illustrates this, showing that lower layers favour local attention while higher layers exhibit long-range token interactions.

This pattern likely reflects the reality of gene interaction at different levels. At the single-nucleotide level, interactions primarily occur between proximal tokens. As we increase the nucleotide length, longer nucleotide segments begin to form regulatory units such as enhancers and silencers \cite{maston2006transcriptional} that can interact at increased distances.

We can interpret the models in more detail by looking at the attention between tokens at different heads. Figure \ref{fig:figure4}(b) further reveals that distinct attention heads at layer 8 capture varying patterns. For instance, head1 primarily captures a 'look-back' pattern evident in the lower triangle attention, head5 seems to focus on long-range interactions, and head8 attends to immediate proximal interactions. Appendix \ref{appendix:c} provides supplementary figures to visualize Attention Weights.

\section{Conclusion and Future Work}
This study presents a task-agnostic hierarchical transformer for one-dimensional data, underscoring the importance of efficient, comprehensive hierarchical models. When applied to genomic assay prediction, Genome Interpreter outperforms conventional models while providing interpretable insights.

As a transformer-based architecture, Genomic Interpreter can be reinforced with pretraining \cite{hendrycks2020pretrained} for improving out-of-distribution prediction and attention flow \cite{abnar2020quantifying} for mapping the obtained attention weights to the original input sequence. While genomic science has emphasized the importance of understanding hierarchical structures, this concept is also critical in other fields, including Natural Language Processing (NLP) where language understanding relies heavily on hierarchical concept interpretation. As a result, 1D-Swin may have the potential to be applied to these fields for capturing long-range, hierarchical information. 


\section*{Acknowledgements}
This work was performed using the Sulis Tier 2 HPC platform hosted by the Scientific Computing Research Technology Platform at the University of Warwick, and the JADE Tier 2 HPC facility. Sulis is funded by EPSRC Grant EP/T022108/1 and the HPC Midlands+ consortium. JADE is funded by EPSRC Grant EP/T022205/1. Zehui Li acknowledges the funding from the UKRI 21EBTA: EB-AI Consortium for Bioengineered Cells \& Systems (AI-4-EB) award, Grant BB/W013770/1. For the purpose of open access, the authors have applied a Creative Commons Attribution (CC BY) licence to any Author Accepted Manuscript version arising.

\bibliography{example_paper}
\bibliographystyle{icml2023}

\newpage
\appendix
\onecolumn
\section{1D-Swin Block}
\label{appendix:a}
The input sequence $\mathbf{x}=[x_0,x_1,...,x_n]$ is partitioned into non-overlapping local windows $\mathbf{w}= [w_0,w_1,...]$, with each window $w_i$ comprising a subset of the sequence $[x_j,x_{j+1},...,x_{j+k-1}]$, given a pre-defined window size $k$. The first Multi-Head Attention (MHA block), as depicted in Figure \ref{fig:architecture}(b), processes each window $w_i$ to generate $w_i'=[x_j',x_{j+1}',x_{j+k-1}']$. This transformation encapsulates local dependencies within each window.

Cross window dependencies are captured through a rolling operation, $R$, which shifts all tokens by a specific distance, $t$. Elements shifted beyond the final position are reintroduced at the start, resulting in a new sequence, $R_t(\mathbf{x'}) = [x_{n-t+1}',...,x_{n}',x_{0}',x_{1}',...,x_{n-t}']$. These shifted elements are reassembled into new windows, denoted as $R(\mathbf{w}')$, where each $R(\mathbf{w}')i$ contains $[x_{j-t}', x_{j+1-t}', x_{j+k-1-t}']$. A second MHA block is then applied to each $R(\mathbf{w}')_i$, after which the inverse rolling operation, $R^{-1}$, is used to restore the original sequence order.

The total computation for each window, illustrated in Figure-\ref{fig:architecture}(b), follows equations \ref{eq:4},\ref{eq:5},\ref{eq:6}:
\begin{equation}
    \label{eq:4}
    \mathbf{w}_i' = f_{\text{MHA}_1}(\mathbf{w}_i)
\end{equation}
\begin{equation}
    \label{eq:5}
    R(\mathbf{w}')_i =  R(f_{\text{concat}}(\mathbf{w}_1',\mathbf{w}_2',...))_i
\end{equation}
\begin{equation}
    \label{eq:6}
    \mathbf{w}''_i = R^{-1}( f_{\text{MHA}_2}(R(\mathbf{w}')_i))
\end{equation}

Given $\mathbf{x} \in \mathbb{R}^{n \times d}$, the spatially reduced token set, $\mathbf{h} \in \mathbb{R}^{\frac{n}{2} \times 2d}$, is computed using a token merging operation, which concatenates pairs of tokens and passes them through a linear layer:

\begin{equation}
    \mathbf{h}  =f_{\text{Linear}}(f_{\text{PairConcat}}(\mathbf{w}'')) = f_{\text{1dSwin}}(\mathbf{x}) 
\end{equation}

\newpage
\section{Architecture Details for Reference Models}

\begin{figure*}[ht]
\vskip 0.1in
\begin{center}
\centerline{\includegraphics[width=\textwidth]{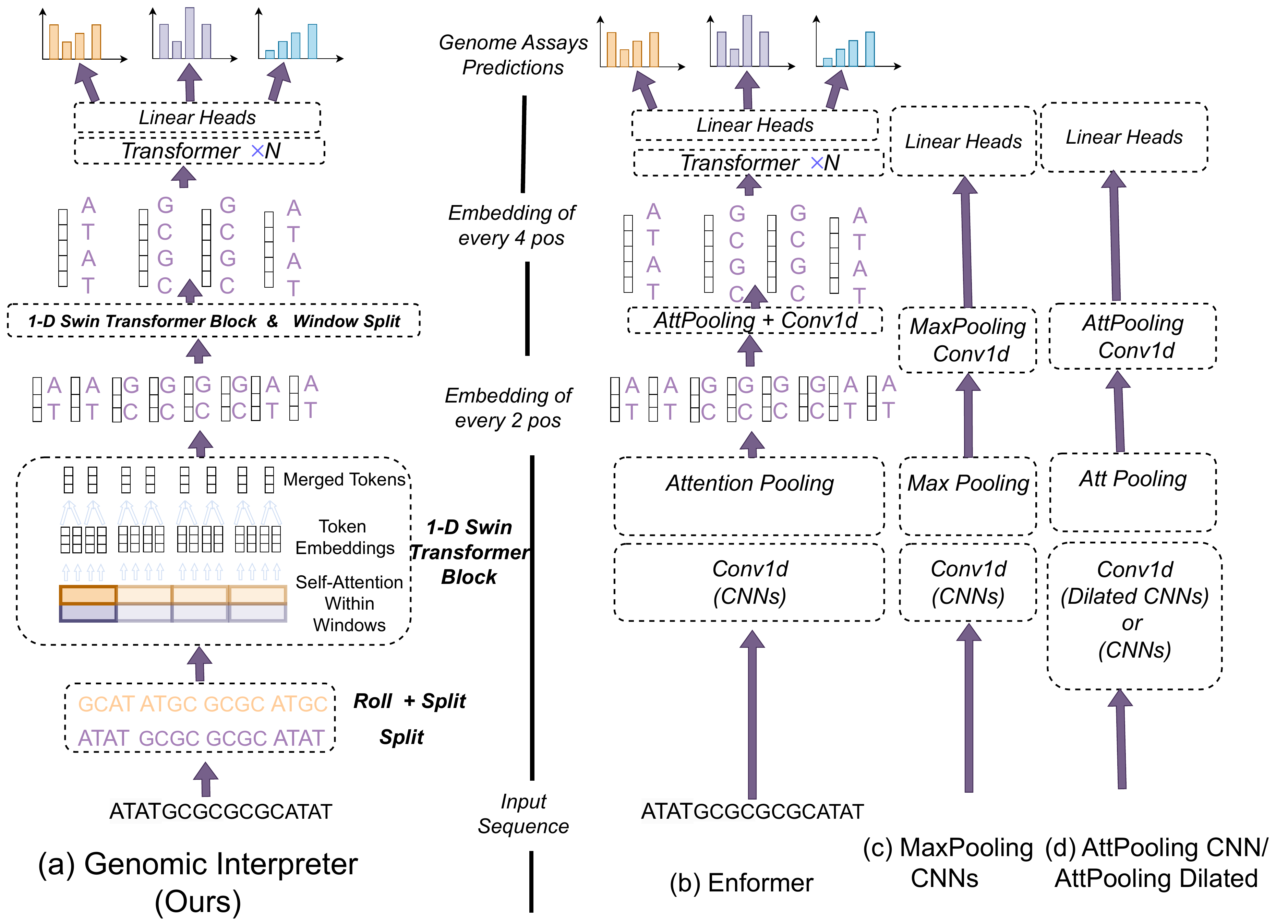}}
\caption{(a) is the architecture of Genomic Interpreter, where no pooling layer is used. (b) shows the architecture of Enformer, which uses attention pooling and CNNs for reducing the spatial dimension and increasing the hidden dimension of each token, finally a transformer layer is appended to aggregate information. (c) Max pooling CNNs use the max pooling layer and remove the transformer layer at the end. (d) Attention pooling CNNs and dilated CNNs use attention pooling layers with normal CNNs or Dilated CNNs.}
\label{app_fig:architecture}
\end{center}
\vskip -0.1in
\end{figure*}

Figure \ref{app_fig:architecture} shows the overall architecture of reference models. Note that Conv1d represents the 1-d convolutional layers with residual links. Relative Positional Embedding is used in Transformer blocks. 
\label{appendix:b}

\newpage
\section{Analysis of Attention Weights of 1D-Swin}
\label{appendix:c}
\subsection{Per-Window View}
 Each layer of 1D-Swin has multiple windows. While  windows at the same layer share the same attention block pairs, these windows output distinct attention weights, highlighting the variety of contextual relations captured within the same layer. The following shows the attention weights obtained at the sixth layer, providing a detailed view across both window and head dimensions.

\begin{figure*}[ht]
\vskip 0.1in
\begin{center}
\centerline{\includegraphics[width=0.7\textwidth]{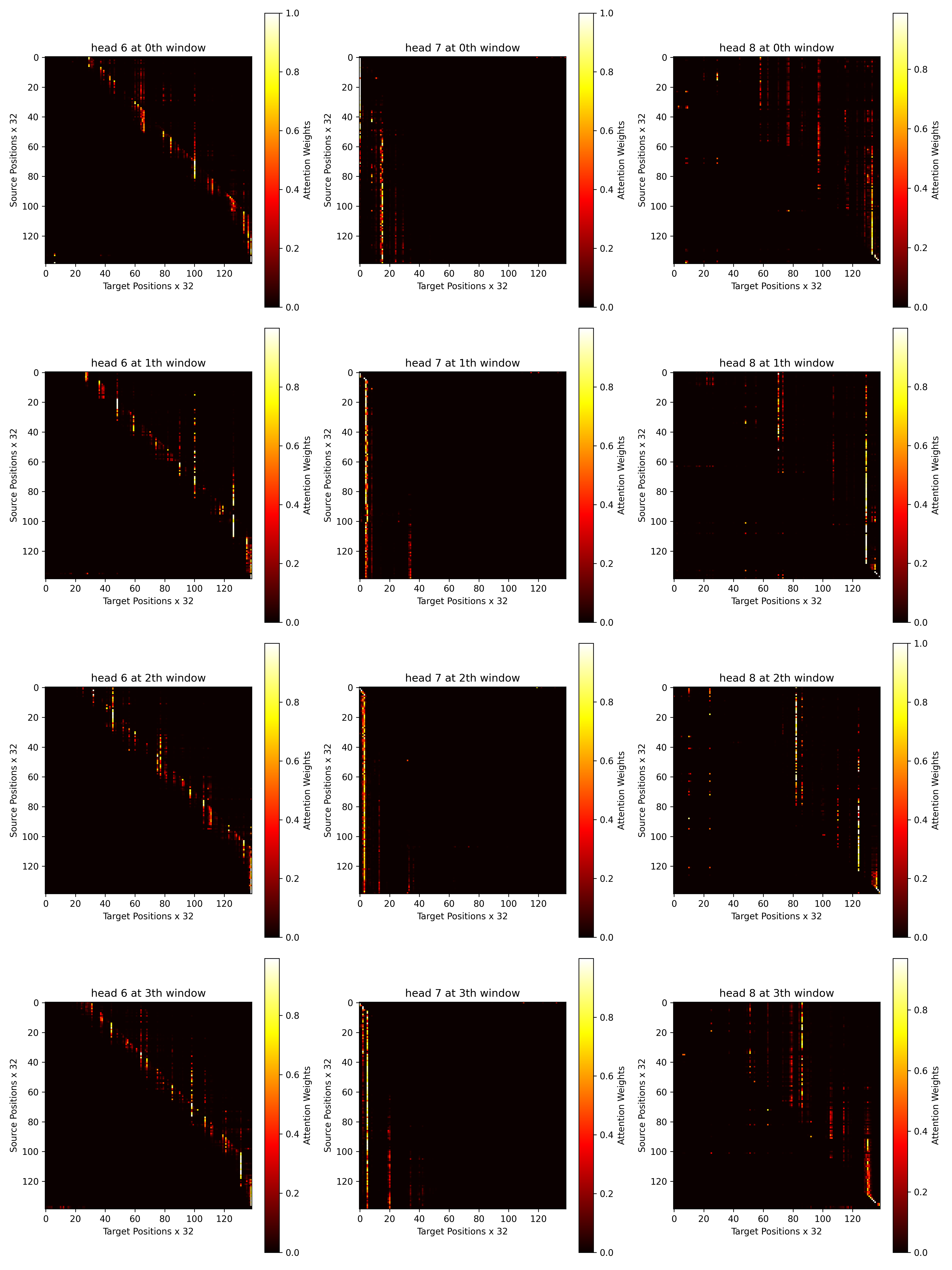}}
\caption{Layer 6 has a window size of 140 tokens, each token corresponds to 32 nucleotides, there are hereby $4=f_{\text{floor}}(\frac{17712}{32*140})$ windows. MHA block pairs share parameters across different windows and are trained to look for certain patterns at different distances, but provide a distinct map providing different sequences within each window.} 
\label{app_fig:att1}
\end{center}
\vskip -0.25in
\end{figure*}

\subsection{Zoom-in View}
The attention weights can also be visualized in a zoom-in view, showing the dependency between tokens in more detail.

\begin{figure*}[ht]
\vskip 0.1in
\begin{center}
\centerline{\includegraphics[width=0.7\textwidth]{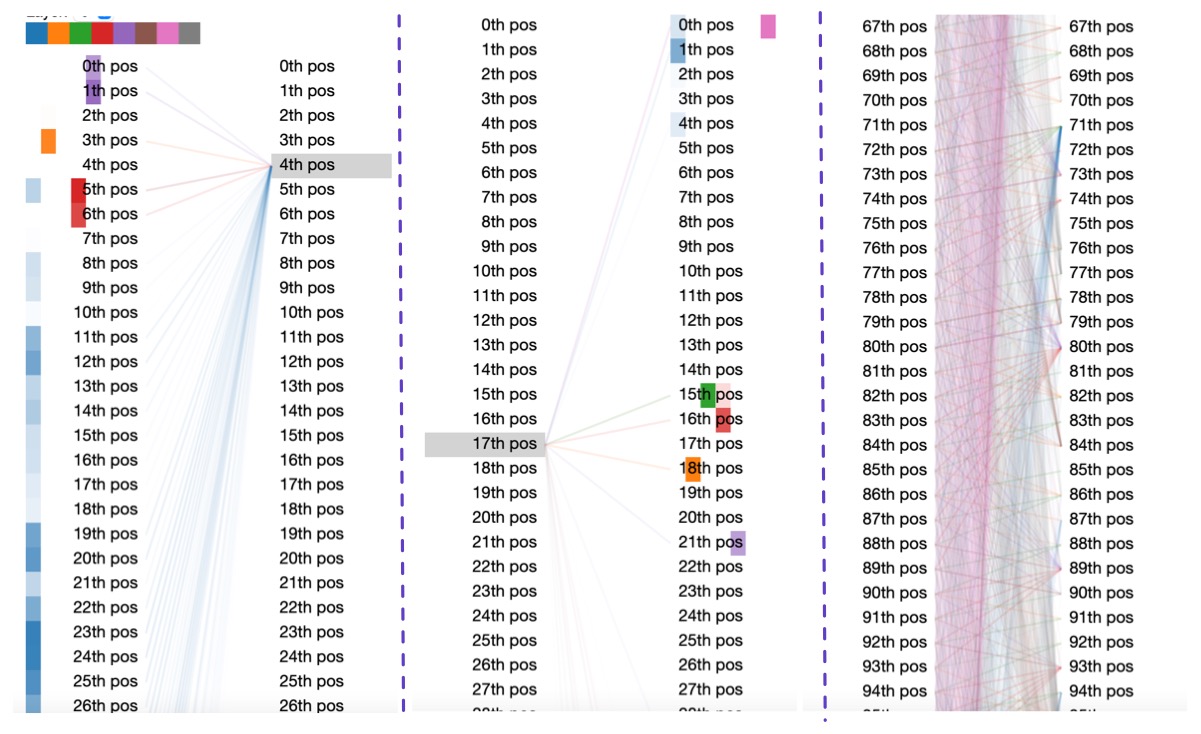}}
\caption{A zoom-in view of how individual tokens attend to each other} 
\label{app_fig:att2}
\end{center}
\vskip -0.25in
\end{figure*}

\section{Training Tricks}
Batch Norm is not used in the model, we apply the layer norm within all the models. This was chosen by considering the potential of numerical instability of batch normalization on the small batches necessitated by GPU memory limitations. Further,  removing batch norm eliminates the need to synchronise the operation across multiple GPUs in the distributed training setting.


\end{document}